\documentclass[aip,jcp,sd,preprint,numerical]{revtex4-1}
\usepackage{float}
\usepackage{xcolor}
\usepackage{ragged2e}
\usepackage{standalone}
\usepackage{amsmath}
\usepackage{gensymb}
\usepackage{epsfig}
\usepackage{setspace}
\usepackage{amstext}
\usepackage{array}
\usepackage{hyperref}
\usepackage{xurl}
\usepackage{graphicx}
\usepackage{multirow}
\usepackage{caption}
\usepackage{subcaption}
\usepackage{adjustbox}
\usepackage{dcolumn}% Align table columns on decimal point
\usepackage{bm}% Bold math

\begin{document}

\title{Naturalization of Text by the Insertion of Pauses and Filler Words}
\author{\textbf{Richa Sharma} 
\\Corresponding Author
\\Bachelors in Technology, Computer Science and Engineering, PES University, Bangalore
\\Address: PES University, 100 Feet Road, BSK III Stage, Bangalore 560085, India
\\Email: richa13sha@gmail.com
\\
}
\author{\textbf{Parth Vipul Shah} 
\\Bachelors in Technology, Computer Science and Engineering, PES University, Bangalore
\\Address: PES University, 100 Feet Road, BSK III Stage, Bangalore 560085, India
\\Email: parthvipulshah@pesu.pes.edu
\\
}
\author{\textbf{Ashwini M. Joshi} 
\\Professor, Computer Science and Engineering, PES University, Bangalore
\\Address: PES University, 100 Feet Road, BSK III Stage, Bangalore 560085, India
\\Email: ashwinimjoshi@pes.edu
\\
}
\date{\today}
\maketitle

\noindent \textbf{Abstract:}
In this article, we introduce a set of methods to naturalize text based on natural human speech. Voice-based interactions provide a natural way of interfacing with electronic systems and are seeing a widespread adaptation of late. These computerized voices can be naturalized to some degree by inserting pauses and filler words at appropriate positions. The first proposed text transformation method uses the frequency of bigrams in the training data to make appropriate insertions in the input sentence. It uses a probability distribution to choose the insertions from a set of all possible insertions. This method is fast and can be included before a Text-To-Speech module. The second method uses a Recurrent Neural Network to predict the next word to be inserted. It confirms the insertions given by the bigram method. Additionally, the degree of naturalization can be controlled in both these methods. On the conduction of a blind survey, we conclude that the output of these text transformation methods is comparable to natural speech.
\\

% \noindent \textbf{Abstract:}
% We study the effect of the coronavirus disease 2019 (COVID-19) in India using the SEIR compartmental model. After it's outbreak in Wuhan, China, it has been imported to India which is a densely populated country. India is fighting against this disease by ensuring nationwide social distancing. We estimate the infection rate to be 0.270 using a least square method with Poisson noise and estimate the reproduction number to be 2.70. We approximate the peak of the epidemic to be August 9, 2020. We estimate that a 25\% drop in infection rate will delay the peak by 11 days for a 1 month intervention period. We estimate that the total individuals infected in India will be approximately 9\% of the total population. 
% \\

\section{Introduction}

% Motivation - 1
Electronic systems that interact with humans are making their way into everyday life. From robots in the house to personal assistants on our phones, these systems aid us in completing both simple and complex tasks. Interfacing with these electronic systems has also evolved. The focus has shifted to voice-based interactions over the more traditional text-based interactions seen in legacy systems\cite{voice}. Voice commands and feedback provide a more natural way for us to interface with these systems. However, even with the rapid development of the state of the art systems, we still largely interact with computerized voices. With this increase in voice-based interaction, we propose a set of methods to naturalize them. Naturalization of speech can be achieved in multiple ways. In this article, we transform text by adding pauses and filler words such as 'uh' and 'um' to an input sentence. Google's Duplex\cite{duplex}, can make calls on your behalf to book appointments or reserve tables and was unveiled in 2018. Despite being a computerized voice, pauses and filler words make the conversation more natural. Motivated by this observation, we propose the following methods. Moreover, these methods can be integrated as a module in a pipeline that takes a voice query as input and outputs a voice response. This module would take the response text from an information retrieval system as input and output naturalized text which can be input for a text-to-speech (TTS) module. \\

% Motivation - 2
Previous cognitive studies of natural human speech indicate that the addition, deletion, modification of words, sentences of standard speech are observed in conversations\cite{cognition}. These can be termed as disfluencies in speech. To achieve naturalization of speech, we resort to the addition of pauses and filler words such as 'uh' and 'um'. These pauses and filler words are an important part of natural conversations and unconsciously end up as a part of them. In natural conversations, pauses and filler words are often used when a conversationalist is either emphasizing on a part of the conversation or gathering their thoughts. By inserting them in standard text, the text is transformed into a more natural text. An example \\

\noindent\fbox{%
    \parbox{\linewidth}{%
       Standard Text: Let's see, Susan is 15. Aundrea is 9. Every stupid cliche you hear about kids, they change your life, they make you a better person, they make you whole, it's all true! Now I get it. \\
       Transformed Text: (um) Let's see, (pause) Susan is 15. Aundrea is 9. every stupid cliche you hear about kids, (pause) they change your life, they make you a better person, they make you whole, (pause) it's all true! Now I get it.
    }%
}
\\

\section{Literature Review}

% Previous works and novelty in this article
As observed, in a conversational context, the transformed text sounds more natural than the standard text. After studying the structure of conversations, we observe the frequency of these filler words in conversations\cite{freq} and the roles they play\cite{disc}. Previously, in \cite{stolcke1} a language model is developed that predicts disfluencies probabilistically and uses an edited, fluent context to predict the following words. In \cite{stolcke2}, a combination of prosodic cues modeled by decision trees and word-based event N-gram language models automatically detects disfluencies. These models detect disfluencies such as pauses and filler words in speech. In \cite{local}, an "Editing Term" is inserted from the speech synthesis inventory and local prosodic modifications are applied at units adjacent to it. This suggests synthetic disfluent speech can be generated by making local changes. In \cite{sundaram}, an empirical text transformation method is proposed for spontaneous speech synthesizers which are trained on transcribed lecture monologue. The novelty of this article includes (a) using transcribed movie dialogues as training data, (b) a hybrid approach in inserting pauses and filler words to increase confidence, (c) tunable degree of naturalization, (d) blind survey to assess the quality of the transformed text.

% Summary of Method
First, we discuss data preprocessing followed by the two proposed methods. The bigram method involves the storing of bigrams with a pause or a filler word and its corresponding frequency in the training data. During transformation, a probabilistic approach is adopted to insert an appropriate pause or filler word. The hybrid approach first transforms text using the bigram method. A sequence to sequence model tuned to the context is then employed to validate these insertions.

\section{Data Preprocessing}

The closest imitation of natural human speech which is well recorded and structured is in the form of dialogues from movie scripts. We use the Movie Dialogue Corpus from Kaggle\cite{cornell} as well as the Baskin Engineering, Film Corpus 2.0\cite{ucsc}. Combined, this forms a corpus of over $2.2 \times 10^{6}$ lines of text. Data preprocessing (I) outputs the training data and the steps involved are
\begin{enumerate}
    \item Extracting the lines containing disfluencies such as pauses and "uh", "um"
    \item Care is taken to identify and extract multiple variations of the above disfluencies such as "uhh" and "uhm"
    \item Two or more periods or dashes are also considered as pauses and are extracted forming an intermediate corpus
    \item Identified pauses are replaced by "(pause)" and "uh", "um" are replaced by "(uh)", "(um)" to maintain uniformity in the final corpus
\end{enumerate}
This forms a corpus of over $2 \times 10^{5}$ lines and serves as the training data.

\section{Methodology and Implementation}
\subsection{Bigrams}
A sentence is taken as an input and the transformed sentence with the inserted pauses and filler words is output. Parameters that control the degree of naturalization can be set. Roman numerals relate to steps in the Figures. Following are the steps of the Bigrams (II) method

\begin{enumerate}
    \item A list of bigrams of words containing pauses and filler words as either the predecessor or the successor to a word is constructed from the training data. The frequency of these bigrams is recorded. Additionally, a special token is appended at the start of each sentence to account for the possibility of an insertion at the beginning of the sentence.
    \item Clean the input sentence by dropping punctuation and numbers. The case is uniformly set to lower. 
    \item Find all possible insertions in the input sentence by splitting it into bigrams and comparing (III) the predecessor and successor with the list of bigrams from step 1. This is the draw set, $D$.
    \item Draw a subset from $D$ based on the degree of naturalization and a probability distribution. The degree of naturalization is said to be the percentage of inserted words given the length of the input sentence.
    \begin{equation}
        \textit{Degree of naturalization} = \frac{\textit{Number of inserted words}}{\textit{Length of input sentence}}
    \end{equation}
    A probability distribution that fits the frequency of bigrams from step 1 is used to construct a subset of $D$. The set of probabilities, $P$ with which a bigram appears in a subset of $D$ is 
    \begin{equation}
        % P(x) = \frac{f(x)}{size(D)}
        P = \{\frac{f(x)}{size(D)}, \forall x \in D\}
    \end{equation}
    where $f(x)$ is the frequency of the bigram in the training data and $size(D)$ is the size of set $D$. The size of the subset of $D$ depends on the degree of naturalization.
    A uniform or any other probability distribution can be used too. The preceding distribution was used in this method as it gave the most intelligible transformations.  
    \item Construct the output sentence by adding the bigrams (filler word included) that are part of the subset of $D$ from step 4 to the input sentence.
\end{enumerate}

It is observed that some bigrams occur more frequently than others in a corpus of text. This is a property of natural speech that we leverage to output more intelligible transformations using the probability distribution mentioned in Equation 2. Step 1 can be performed independently and pre-computed before the method sees an input sentence. Therefore, this step does not contribute to the overall speed of the transformation. Table \ref{tab1} lists the five most frequently occurring bigrams from the training data.

\begin{table}[H]
\caption{Most frequently occurring bigrams and their frequencies in the training data.}
\label{tab1}
\begin{ruledtabular}
\begin{tabular}{l l c}
Predecessor& 
Successor& 
Frequency \\
\hline
(pause) & I & 13062\\
(pause) & and & 8364\\
you & (pause) & 4533\\
and & (uh) & 100\\
(uh) & I & 88\\
\end{tabular}
\end{ruledtabular}
\label{tab1}
\end{table}

A valid subset of $D$ may not be possible if the training data has no occurrence of the bigrams present in the input sentence. Therefore, we propose a Fallback method (IV) based on Parts of Speech (POS) tagging\cite{pos}. This method is a variation of the method proposed above. A layer of abstraction is added by which the POS tag of the training data and input sentence are considered for constructing the bigrams. However, this leads to a significant delay in transformation as the input sentence is first tagged.

\subsection{Hybrid}

Instead of probabilistically choosing a subset of $D$, in this method, the insertions are confirmed using a Recurrent Neural Network (RNN). The RNN predicts the next word based on substrings of the input sentence, a seed phrase. If the predicted word matches a filler word, it is inserted in the output sentence. The RNN in the training (V) phase uses the training data to learn sequences. To train the model

\begin{enumerate}
    \item Pre-process the training data. 500 of the most representative sentences of the corpus are selected as training data.
    \item Instantiate a tokenizer. The tokenizer converts the input data text into sequences. The sequence length is set to three. The method uses the context of the two preceding words to predict one word.  
    \item The sequences are converted to categorical data and an RNN is trained using parameters defined in Table \ref{tab4}.
\end{enumerate}

\begin{table}[H]
\caption{Architecture and Training Parameters of the Recurrent Neural Network (RNN).}
\label{tab4}
\begin{ruledtabular}
\begin{tabular}{l l}
Parameter& 
Value \\
\hline
Layers &  1 \\
Units B & 100 LSTM\cite{lstm} \\
Loss Function &  Categorical Cross-entropy\cite{cce} \\
Optimizer &  Adam\cite{adam} \\
Epochs & 300 \\
\end{tabular}
\end{ruledtabular}
\label{tab4}
\end{table}

We use a simple model as this ensures transformation speed is high. The training data consists of only a subset of the corpus because, for larger training data, a drop in accuracy is observed as the simple model fails to fit the large sample space. A sequence length of three was found to give the highest accuracy. This simple model can be improved to obtain more intelligible transformations.
To transform an input sentence after training is complete

\begin{enumerate}
    \item Clean the input sentence by dropping punctuation and numbers. The case is uniformly set to lower.
    \item From the Bigrams (VI) method, obtain the draw set, $D$. However, discard all the bigrams that have a filler word as the predecessor.
    \item A substring of the input sentence constructed up to an insertion from $D$ is passed through the model to predict the next word. 
    \item Following this, a query (VII) is made. If the next word is the filler word from the bigram, the insertion is confirmed. This is the intermediate output. Based on the degree of naturalization, a subset of all confirmed insertions are selected and the output sentence is constructed.
\end{enumerate}

Sample incorrect and correct predictions for an input sentence "Let us try this one more time." follow \\
\\
\noindent\fbox{%
    \parbox{\linewidth}{%
        Substring of the input sentence: Let \\
        Bigram from $D$: let, \textit{(pause)} \\
        Hybrid predicted next word: \textit{me}\\
        \textbf{Incorrect Prediction} \\
        Intermediate output: Let us try this one more time
    }%
}
\\ 
\noindent\fbox{%
    \parbox{\linewidth}{%
        Substring of the input sentence: Let us\\
        Bigram from $D$: us, \textit{(pause)} \\
        Hybrid predicted next word: \textit{(pause)}\\
        \textbf{Correct Prediction} \\
        Intermediate output: Let us (pause) try this one more time
    }%
}
\\

\begin{figure}[H]
\centerline{\includegraphics[width=\columnwidth]{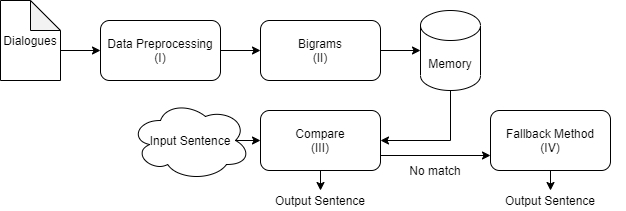}}
\caption{High level architecture of the bigram text transformation method. Roman numerals relate to steps in the text.}
\label{fig3}
\end{figure}

\begin{figure}[H]
\centerline{\includegraphics[width=\columnwidth]{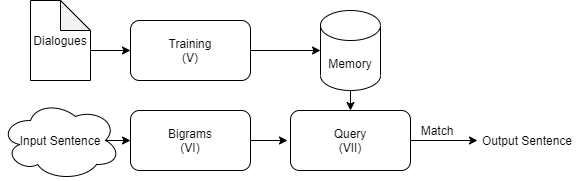}}
\caption{High level architecture of the hybrid text transformation method. Roman numerals relate to steps in the text.}
\label{fig4}
\end{figure}

We observe that the model confirms a pause more frequently than the other filler words. According to \cite{cognition}, pauses in natural language can be replaced by other filler words. Longer sentences are also more likely to begin with these filler words. Therefore, we take the liberty to replace the first $n$ pauses by the other filler words where $n$ is based on the degree of naturalization. Figure \ref{fig3} and \ref{fig4} details the high level architecture of the bigram and hybrid methods.

% \begin{figure}[H]
%     \centering
%     \includegraphics[scale=0.7]{new_cases.jpg}
%     \caption{\label{new_cases}Daily increase in confirmed cases of COVID-19 in India. Day 0 is January 25, 2020 and day 92 is April 26, 2020. Data is taken from Center for Systems Science and Engineering (CSSE) at Johns Hopkins University (JHU).\cite{github}}
% \end{figure}

After detailing the proposed methods, we state that richer training data is a sure way to output more intelligible transformations because the proposed methods rely heavily on the training data to generate possible insertions. Richer training data can be obtained by including a range of contexts if the input sentences are going to be general-purpose. Training data can include specific contexts and introduce bias if the input sentences are going to be from a single context. Additionally, if the training data is a close representation of natural speech with the correct disfluencies inserted at the correct positions, the methods improve. 

\section{Results and Discussion}

As there are multiple ways of achieving natural speech, we aim to transform the text into the most natural-sounding version. To illustrate this, following is an example
\\

\noindent\fbox{%
    \parbox{\linewidth}{%
       Variation 1: This is (pause) the first time I am (uh) coming here. \\
       Variation 2: This is the (uh) first time (pause) I am coming here. \\
       Variation 3: This (uh) is the first time I am coming (pause) here.
    }%
}
\\

Therefore, we use a medium-sized pool of 275 odd individuals to evaluate the outputs and determine if they sound natural or not. We refer to them as respondents hereafter. The outputs are generated by both, the bigram and the hybrid method. Degree of naturalization is 9.5\% for both methods. While answering the question | 'Which of the following sounds more natural?', judging between two text excerpts is more difficult than judging between two voice clips of the same text excerpts. Hence, the respondents were presented with two voice clips. One was a performance of an original interview's transcript and the other was a performance of the transformed text of the same interview's transcript. Both were performed by the same voice actor. Here is an example of the output text from the bigram method followed by an example from the hybrid method
\\

\noindent\fbox{%
    \parbox{\linewidth}{%
       Real Interview: When we actually started recording the album we had this beautiful place when we like rented this kind of beach (pause) shack and (um) that's the only thing I asked for in the budget though.\cite{kevin}\\
       Generated: When we actually started recording the (pause) album we had this beautiful place when we like rented this kind of beach shack (uh) and that's the (pause) only thing I asked for in the (pause) budget though.

    }%
}
\\

\noindent\fbox{%
    \parbox{\linewidth}{%
       Real Interview: I spend a week in a year where I just go off and (uh) read (pause) people's PhD theses and new things that are going on in the field.\cite{bill}\\
       Generated: I spend a week in a year where I just (uh) go off and read people's PhD theses and new things (uh) that are (pause) going on in the field.
    }%
}
\\

\begin{figure}[H]
\centerline{\includegraphics[width=\columnwidth]{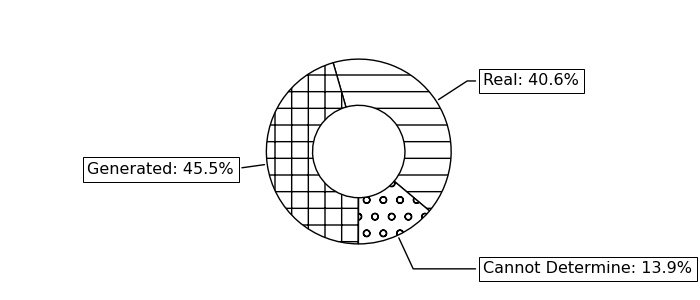}}
\caption{Percentage of respondents determining the more natural sounding voice clip. Transformation method is bigram.}
\label{fig1}
\end{figure}

\begin{figure}[H]
\centerline{\includegraphics[width=\columnwidth]{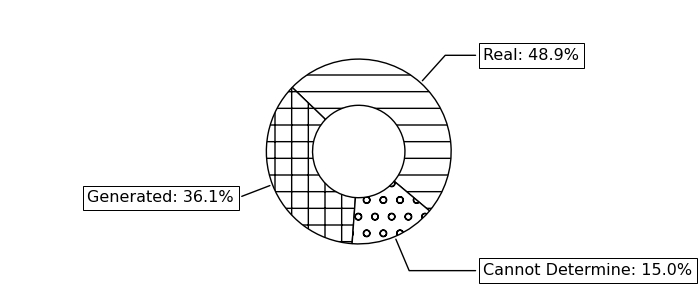}}
\caption{Percentage of respondents determining the more natural sounding voice clip. Transformation method is hybrid.}
\label{fig2}
\end{figure}

A "Cannot Determine" choice was also included as this would indicate near-natural transformation by our methods. This was repeated twice per method with different interviews to test it's robustness when the context of the input text was varied. Table \ref{tab2} and \ref{tab3} list results of each sample, each method. Figures \ref{fig1} and \ref{fig2} illustrate only the summary for each method.

\begin{table}[H]
\caption{Percentage of respondents determining the more natural sounding voice clip. Transformation method is bigram.}
\label{tab2}
\begin{ruledtabular}
\begin{tabular}{c c c c}
Sample& 
Real& 
Generated&
Cannot Determine \\
\hline
% Sample A & 84 & 138 & 54 \\ % total: 276
% Sample B & 140 & 113 & 23 \\ % total: 276
A & 30.4\% & 50.0\% & 19.6\% \\
B & 50.7\% & 40.9\% & 8.3\% \\ 
\end{tabular}
\end{ruledtabular}
\label{tab2}
\end{table}

\begin{table}[H]
\caption{Percentage of respondents determining the more natural sounding voice clip. Transformation method is hybrid.}
\label{tab3}
\begin{ruledtabular}
\begin{tabular}{c c c c}
Sample& 
Real& 
Generated&
Cannot Determine \\
\hline
% Sample C & 144 & 99 & 33 \\ % total: 276
% Sample D & 126 & 100 & 50 \\ % total: 276
C & 52.2\% & 35.9\% & 12.0\% \\
D & 45.7\% & 36.2\% & 18.1\% \\ 
\end{tabular}
\end{ruledtabular}
\label{tab3}
\end{table}

If "Generated" and "Cannot Determine" together form the majority of responses ($\geq 50\%$), our methods can be said to effectively naturalize text. The "Cannot Determine" choice indicates that the transformation was indiscernible from the original text. Counter-intuitively, the hybrid method performs not as well as the bigram method. However, this could be a result of the bias concerning the context of the sample interviews. The methods may perform poorly in certain contexts. We reiterate that the methods are highly sensitive to the context of the training data. As the size of the training data and the breadth of the context increase, the methods get more robust and consequently, the transformations more natural. \\

As stated previously, our methods of text transformation are fast. This is to ensure that this text transformation module can be integrated before a TTS module. Table \ref{tab5} details the timing data in seconds for both methods and their averages. All runs are performed with equal computing power to maintain uniformity. All samples are long sentences and the degree of naturalization is 9.5\% for both methods. Sentence length and degree of naturalization are directly proportional to the time taken to generate an output sentence. We observe that the bigram method is multiple factors faster than the hybrid method.

% TODO: add justification for timing data and timing data

\begin{table}
\caption{Timing Data for the bigram and hybrid methods.}
\label{tab5}
\begin{ruledtabular}
\begin{tabular}{c c c}
Sample& 
Bigram Time (s)& 
Hybrid Time (s) \\
\hline
E & 0.236 & 3.649 \\
F & 0.399 & 5.835 \\
G & 0.213 & 2.646 \\
H & 0.574 & 5.270 \\
I & 0.358 & 4.624 \\
\hline
Average & 0.356 & 4.405 \\

\end{tabular}
\end{ruledtabular}
\label{tab5}
\end{table}

% \def\arraystretch{1.5}
% \begin{table}[H]
% \caption{\label{tab:rate}
% Growth Factor, $r$}
% \begin{ruledtabular}
% \begin{tabular}{l c c}
% Date &No. of confirmed cases &Growth factor, $r$ \\
% \hline
% April 22, 2020 &1290 &- \\
% April 23, 2020 &1707 &1.323 \\
% April 24, 2020 &1453 &0.851 \\
% April 25, 2020 &1753 &1.206 \\
% \end{tabular}
% \end{ruledtabular}
% \end{table}

\section{Conclusion}

In this article, we propose a novel, fast transformation method that has tunable naturalization. It also uses a hybrid approach to confirm the transformations. We evaluate our methods using a medium-sized pool of individuals to evaluate the outputs and determine if they are natural. We conclude that our methods yield near-natural transformations of text and can be integrated as a module into a voice query to voice response pipeline.

\medskip

\noindent \textbf{Acknowledgements:} The authors would like to thank PES University and Dr. S S Shylaja, Chairperson, Department of Computer Science and Engineering, PES University for their support. 
\\
\smallskip

\noindent \textbf{Conflicts of Interests:} The author declares no conflict of interest
\\
\smallskip

\noindent \textbf{Declarations of Interest:} None
\\
\smallskip

\noindent \textbf{Funding:} None

\medskip

% Hybrid approach for confirmation 
% Fast lookup for insertion of filler words
% Tunable naturalization
% Survey conducted to evaluate the accuracy of generated speech


\begin{thebibliography}{99}
\bibitem{voice} Voice Assistants: How Artificial Intelligence Assistants Are Changing Our Lives Every Day
\url{https://www.smartsheet.com/voice-assistants-artificial-intelligence} Accessed: January 3, 2020

\bibitem{duplex} Google Duplex: An AI System for Accomplishing Real-World Tasks Over the Phone. \url{https://ai.googleblog.com/2018/05/duplex-ai-system-for-natural-conversation.html} Accessed: January 3, 2020

\bibitem{cognition} Clark, Herbert \& Fox Tree, Jean. (2002). Using uh and um in spontaneous dialog. Cognition. 84. 73-111. 10.1016/S0010-0277(02)00017-3.

\bibitem{freq} Yang Liu, E. Shriberg, A. Stolcke, D. Hillard, M. Ostendorf and M. Harper, "Enriching speech recognition with automatic detection of sentence boundaries and disfluencies," in IEEE Transactions on Audio, Speech, and Language Processing, vol. 14, no. 5, pp. 1526-1540, Sept. 2006, doi: 10.1109/TASL.2006.878255.

\bibitem{disc} Julia Hirschberg and Diane Litman. 1993. Empirical studies on the disambiguation of cue phrases. Comput. Linguist. 19, 3 (September 1993), 501–530.

\bibitem{stolcke1} A. Stolcke and E. Shriberg, "Statistical language modeling for speech disfluencies," 1996 IEEE International Conference on Acoustics, Speech, and Signal Processing Conference Proceedings, Atlanta, GA, USA, 1996, pp. 405-408 vol. 1, DOI: 10.1109/ICASSP.1996.541118.

\bibitem{stolcke2} Stolcke, Andreas \& Shriberg, Elizabeth \& Bates, Rebecca \& Ostendorf, Mari \& Hakkani-Tur, Dilek \& Plauche, Madelaine \& Tur, Gokhan \& Lu, Yu. (1998). Automatic detection of sentence boundaries and disfluencies based on recognized words. 

\bibitem{local} On the generation of synthetic disfluent speech: local prosodic modifications caused by the insertion of editing terms. J Adell, A Bonafonte, DE Mancebo - INTERSPEECH, 2008

\bibitem{sundaram} Sundaram, Shiva \& Narayanan, Shrikanth (2003). An empirical text transformation method for spontaneous speech synthesizers.

\bibitem{cornell} Movie Dialog Corpus \url{https://www.kaggle.com/Cornell-University/movie-dialog-corpus} Accessed: January 30, 2020

\bibitem{ucsc} Marilyn A. Walker, Grace I. Lin, Jennifer E. Sawyer. "An Annotated Corpus of Film Dialogue for Learning and Characterizing Character Style." In Proceedings of the 8th International Conference on Language Resources and Evaluation (LREC), Istanbul, Turkey, 2012.

\bibitem{pos} Bird, Steven, Edward Loper and Ewan Klein (2009). Natural Language Processing with Python. O’Reilly Media Inc.

\bibitem{lstm} Hochreiter, Sepp \& Schmidhuber, Jürgen. (1997). Long Short-term Memory. Neural computation. 9. 1735-80. 10.1162/neco.1997.9.8.1735. 

\bibitem{cce} Categorical Cross Entropy Loss Function
\url{https://peltarion.com/knowledge-center/documentation/modeling-view/build-an-ai-model/loss-functions/categorical-crossentropy}
Accessed: February 2, 2020

\bibitem{adam}	Diederik P. Kingma, Jimmy Ba, Adam: A Method for Stochastic Optimization, arXiv:1412.6980 [cs.LG]

\bibitem{kevin} Tame Impala: ‘The Slow Rush’ Full Interview | Apple Music
\url{https://www.youtube.com/watch?v=X3RrKNOT3Ws} Accessed: March 14, 2020

\bibitem{bill} 1991 Interview with Bill Gates
\url{https://www.youtube.com/watch?v=6V6Gir1Dyfs} Accessed: March 14, 2020

\end{thebibliography}
\end{document}